\documentclass[conference]{IEEEtran}
\ifCLASSINFOpdf
\else
\fi
\usepackage{svg}
\usepackage{lineno}
\usepackage{graphicx}
\usepackage{soul}
\usepackage{todonotes}
\usepackage[utf8]{inputenc}
\usepackage{url}
\usepackage{float}
\usepackage{afterpage}
\usepackage{placeins}
\usepackage{units}
\usepackage{amsmath}
\usepackage{multirow}
\usepackage{listings}
\usepackage{color, colortbl}
\definecolor{LightGray}{rgb}{0.95,0.95,0.95}
\definecolor{LightCyan}{rgb}{0.9,1,1}
\definecolor{LightGreen}{rgb}{0.9,1,0.9}
\definecolor{LightRed}{rgb}{1,0.9,0.9}
\definecolor{Red}{rgb}{0.2,0,0}
\definecolor{Green}{rgb}{0,0.2,0}
\begin{document}

%
\title{A Human-in-the-Loop Approach based on Explainability to Improve NTL Detection}

\author{\IEEEauthorblockN{Bernat Coma-Puig}
\IEEEauthorblockA{Barcelona School of Informatics\\
Universitat Politecnica de Catalunya\\
08034 Barcelona, Spain\\
Email: bcoma@cs.upc.edu}
\and
\IEEEauthorblockN{Josep Carmona}
\IEEEauthorblockA{Barcelona School of Informatics\\
Universitat Politecnica de Catalunya\\
08034 Barcelona, Spain\\
Email: jcarmona@cs.upc.edu}
}


%


\maketitle

\begin{abstract}
Implementing systems based on Machine Learning to detect fraud and other Non-Technical Losses (NTL) is challenging: the data available is biased, and the algorithms currently used are black-boxes that cannot be either easily trusted or understood by stakeholders. This work explains our human-in-the-loop approach to mitigate these problems in a real system that uses a supervised model to detect Non-Technical Losses (NTL) for an international utility company from Spain. This approach exploits human knowledge (e.g. from the data scientists or the company's stakeholders) and the information provided by explanatory methods to guide the system during the training process. This simple, efficient method that can be easily implemented in other industrial projects is tested in a real dataset and the results show that the derived prediction model is better in terms of accuracy, interpretability, robustness and flexibility. 
\end{abstract}


%
\IEEEpeerreviewmaketitle

\section{Introduction}
\label{sec:Introduction}

Historically NTL detection systems were rule-based and aimed to emulate human knowledge through a few rules. However, the current availability of large amounts of data has led modern fraud detection systems to use machine learning techniques.  With this change, the systems become smarter (e.g. they can learn more complex patterns or can be updated more easily) but also more difficult to understand, especially when black-box algorithms are used to make predictions. 

An example of an NTL detection system that uses a black-box algorithm (a
Gradient Boosting Decision Tree) is the system that the Universitat Politecnica de Catalunya has built for an international
utility company from Spain. Our system \cite{mdpi,coma2016fraud} has achieved
good results. However, the
system had problems in terms of fairness
and robustness because the data come from observational data produced for other purposes, and therefore the available information does not reliably represent  reality (i.e. is biased). The assumption of i.i.d. (independent and identically distributed data) between the labelled information and reality (i.e. the company's customers) is not met and, therefore, the desired characteristics of reliability and fairness in our system were not fully achieved. These problems, as we explain in \cite{ExpertApp}, were partially solved with the introduction of Shapley Values \cite{shapley1953value} from SHAP~\cite{shap} since we started to understand the deficiencies of our system and validate our work beyond benchmarking (e.g. by verifying the patterns learned by the model). However, the fact that our system is implemented in different domains did not make it possible to implement generic solutions to solve all the existing problems, as the biases derived from the observational data translated differently in each domain and time of the campaign.
 
In this work, we present our step forward to exploit the information provided by the Shapley Values: to convert the process of building our model into a human-in-the-loop process controlled by the stakeholder in charge of the NTL detection process. In each iteration, this specialist analyses what the model has learned and implements feature engineering to improve the model if it detects an undesired pattern, a bias, or an unused feature. After several iterations, as we exemplify in this work, the resulting model is better in terms of accuracy, robustness, interpretability, generalizability, flexibility, and simplicity.
 
The paper is organized as follows. Section \ref{sec:RW} contextualizes the work done and explains the existing literature. Section \ref{sec:ExpNTL} explains our case study, briefly summarizing how it works and its challenges, and proposes the human-in-the-loop approach to exploit the information from the Shapley Values. In Section \ref{sec:CaseStudy} we test this approach in a real dataset. We conclude this work with Section \ref{sec:Conclusions}, summarizing the benefits of our approach and introducing possible future work.

\section{Related Work and Preliminaries}
\label{sec:RW}
\subsection{Related Work in NTL detection}

The use of Supervised Machine Learning techniques (especially black-box algorithms) to detect NTL is extensive in the literature. As our system implements a CatBoost \cite{catboost} predictive model to detect NTL in Spain, we would like to highlight \cite{Endesa}, a similar approach to ours (it uses an XGBoost \cite{xgboost} model and is also implemented in Spain). It is also common to see examples of using Support Vector Machine (SVM) in the literature. Two examples of this are \cite{SVM_1}, which uses a radial basis function as kernel, and \cite{depuru2013high}, which uses a sigmoid kernel. The popularity of Artificial Neural Networks can also be seen in NTL literature: \cite{Ann-Based,ann_brasil,ann_2,nizar2008power} are four examples of it. Finally, we should mention \cite{OPF_1,OPF_2}, two examples of using Optimal-Path Forest Classifier \cite{papa2009supervised} to detect NTL, a rather new non-parametric technique that is grounded on partitioning a graph into optimum-path trees. 

The classical approach of using rule-based systems that aim to reproduce human expert knowledge is becoming obsolete due to the difficulty of exploiting a large amount of data, but many approaches in the literature still combine complex Supervised methods with Rule systems. For instance, in \cite{SVM+FR} the approach proposed in \cite{SVM_1} is improved by introducing Fuzzy-Rules. Similarly, in \cite{depuru2013high} a Rule Engine is combined with a Sigmoid SVM to maximize the predictive capacity of the system. In \cite{guerrero2014improving} text mining, neural networks, and statistical techniques are combined to build a rule-based system.

The extensive NTL literature includes many examples with unsupervised methods.  In \cite{clust_1} and \cite{clust_2} there are two examples of using clustering; and in \cite{cabral2008fraud} there is an example of using unsupervised neural networks (Self-Organizing Maps). In \cite{spiric2015fraud} and \cite{liu2015cyberthreat} there are two other examples of unsupervised methods that focus on statistical control to detect NTL cases. 

In addition to the data-oriented solutions explained above, other approaches exploit the existence of sensors and the smart meters' capacities. In \cite{loadflows} NTL is detected by analyzing the load flow; and in \cite{xiao2013exploring} a structure is presented where an \emph{inspector} meter controls the customers' meters, a popular approach that facilitates NTL detection in highly populated cities. Different examples of network-oriented NTL detection, data-oriented and even hybrid methods (that combine both approaches) can be seen in \cite{messinis2018review}.

\subsection{Accuracy, Interpretability, Explainability and Human Knowledge}

The use of these black-box algorithms is raising some concerns in the AI community since this lack of transparency prevents extensive use of these techniques in certain cases, e.g. in medicine (accountability issues if the predictive algorithm has a mistake) or in justice (where the predictive algorithms can learn biases such as racism from judges). In other domains, the lack of transparency may not lead to these ethical concerns, but it may make it difficult to achieve a robust model that learns causal patterns without bias. 

Explainable AI (XAI) \cite{gunning2017explainable} is a new trend in Artificial Intelligence that proposes the use of methods to guarantee explainable and very accurate models that stakeholders can humanly understand. We highlight two model-agnostic ad-hoc methods tested in our system that provide explainability to the black-box algorithms, LIME \cite{lime} (an approach that builds an interpretable model $L$ that surrogates the complex model $M$ for an instance $x$), and SHAP \cite{shap}, an approach based on Shapley Values \cite{shapley1953value}, a game theory approach to fairly distribute the payout among the players that have collaborated in a cooperative game. The SHAP library adapts this idea\footnote{SHAP considers the payoff as the prediction and the values of the features the players of the cooperative game.} to determine how the values of the features of an instance $x$ influenced the prediction made for the supervised model $M(x)$. It is usually defined as follows: 

\begin{equation*}
\resizebox{\columnwidth}{!}{$
   \psi_i=\sum_{S\subseteq\{ x_1, \ldots, x_m \}\setminus\{x_i\}}\frac{|S|!\left(p-|S|-1\right)!}{p!}\left(val\left(S\cup\{x_i\}\right)-val(S)\right)
   $}
\end{equation*}

In the equation, variable $S$ runs over all possible subsets of feature values, the term $val\left(S\cup\{x_i\}\right)-val(S)$ is the marginal value of adding $x_i$ in the prediction using only the set of feature values in $S$, and the term $\frac{|S|!\left(p-|S|-1\right)!}{p!}$ corresponds to the permutations that can be made with subset size $|S|$, to weight different sets differently in the formula. All possible subsets of attributes are considered, and the corresponding effect is used to compute the Shapley Value of $x_i$. In our system we use the Tree Explainer to extract the Shapley Values, the specific method from SHAP for Tree Models \cite{lundberg2018consistent}.

The promise of the explanatory algorithms is that the discussion about the trade-off between accuracy vs interpretability, i.e. the necessity of deciding between using complex algorithms (e.g. Deep Learning, Ensemble Trees or Support Vector Machines) or interpretable algorithms like Linear Regression or Decision Trees vanishes since several goals \cite{arrieta2020explainable} can be achieved through the explanatory algorithms (i.e. trustworthiness, causality, transferability, informativeness, confidence, fairness, accessibility, interactivity and privacy awareness). However, relevant work (e.g. \cite{rudin2019stop}) still advocate using interpretable algorithms and consider that many of these goals that in a great deal of cases are easily achievable through the interpretable models cannot be easily achieved using black-box algorithms.

In any event, the need for humanly validating a model to avoid biases and other problems calls into question the classical approach of benchmarking an intelligent system as the \emph{skill} of correctly doing a specific task, e.g. a predictive model that assigns a label \cite{chollet2019measure,drummond2010warning}, masking what should define intelligence in artificial intelligence, e.g. the ability of generalizing what the system learns, and how it has to be benchmarked, i.e. against human intelligence.

All the literature referred to in this Section, combined with other more classical machine learning techniques that share some similitude with the approach proposed in this work (e.g. feature selection \cite{dash1997feature}, active learning \cite{settles2009active} and human-in-the-loop \cite{zanzotto2019human}), inspired us in the development of our proposal.





\section{Our approach in the NTL detection system}
\label{sec:ExpNTL}

\subsection{The Supervised Approach}

Our initial approach \cite{coma2016fraud,mdpi} was a supervised binary approach summarized as follows. First, the Stakeholders delimited the segmentation of the campaign (the type of utility, region and tariff) and the system profiled the visited customers in the past (i.e. the NTL and non-NTL cases), and in the present (the customers to be predicted). Then, a model is trained with the historical information and the profiles at present got a binary prediction. Finally, the top-scored customers were included in a report. Initially, this report was analyzed, and the company decided which customers were to be visited.


To profile the customer we included several features, summarized as follows:
\paragraph{Consumption Features} From the consumption data available, we build numeric features that refer to the kWh consumed by (or billed to) the customer during a period, information regarding the difference between the customer's consumption in two distinct periods (to detect abrupt consumption drop), ratios between the consumption of the customer against other similar customers in the same period (to detect long periods of abnormal consumption behavior), as well as the customer’s consumption curve to detect abnormal consumption peaks. 

\paragraph{Visit features} Most of the features that can be extracted from the visits made by the technicians to the customers are very important for deriving a supervised problem. \textit{Labelled instances}: When we profile the customers in month m, all the visits performed in that month are the labelled instances for the supervised training stage. \textit{Visit information:} A visit in month $m$ is the label of the profile from month $m$, but becomes a feature when profiling $m+t$, $t>=1$. 

\paragraph{Static features}

The static information is used to segment the customers in different domains (e.g., the tariff) and generate additional features.  

\paragraph{Sociological features}

The aim of including sociological and geographical information is to nuance the customer's final score; for instance, if we accept the premise that in poorer regions, the people may commit more fraud, the system should prioritize the abnormal behaviors from regions with lower incomes.

A more extensive explanation of the supervised system and the variables is available in \cite{mdpi,coma2016fraud, ExpertApp}.

\subsection{Bypassing the lack of robustness and interpretability}
\label{sec:biasesInt}

Despite the good results achieved in our system\footnote{For instance, for customers with a cancelled contract, our system has achieved many campaigns of around 50\% of precision.}, it lacked robustness, not achieving very successful campaigns consistently. In part, this could be justified by the difficulty of detecting fraud in certain domains due to the lack of labelled information or the very low proportion of fraud that may exist in a developed country. However, the use of observational data also hindered the success of our system since the labelled information is not an 'independent and identically distributed' sample from the company's customers. Instead, the labelled information over-represents certain types of customers (e.g. recidivist fraud customers) and under-represents others (e.g. customers from regions where the company did not consider it interesting to generate campaigns due to business-related considerations). Therefore, there exist biases and dataset-shift\footnote{Dataset-shift occurs when the distribution of the training dataset and the test dataset differ, making it difficult to train robust models.}. In \cite{mdpi} we explain how these problems were partially mitigated by implementing a segmentation of the customers. However,  more technical approaches were tested (e.g., applying weights to the under-represented customers) with inconclusive results. 

In many cases, the use of black-box algorithms blinded us to provide solutions to mitigate the problems in robustness. However, in \cite{ExpertApp} we explain how we bypassed the lack of transparency by using the Shapley Values \cite{shapley1953value} from SHAP \cite{shap} to take well-reasoned algorithmic decisions beyond benchmarking\footnote{Our labelled information is biased and, therefore, the classical approach of sampling a validation dataset to analyze the generalizability cannot be fully trusted.}. For instance, thanks to explanatory algorithms, we detected that a regression approach in which the target to predict is the energy to recover in the NTL provides better results than the classical classification approach, recovering more energy but especially learning more reliable patterns that would generalize better on unseen data. Thus, the introduction of interpretability helped us to implement good solutions that would improve the predictions of our system in any domain.

\subsection{Mitigating the existing problems for each model built}

\subsubsection{The problem}

Although explainability gave us the possibility of improving overall our system, there are still specific problems that cannot be generically solved in our system. In other words, the use of observational data that is constantly changing\footnote{As explained in Section \ref{sec:ExpNTL}, the campaigns' results were included in the system as new labelled information.} translates differently in each model build and needs specific solutions. Despite this, these biases would be mostly easily detectable for specialized stakeholders, as now the system is more transparent thanks to the in-depth explanation of the model with Shapley Values. However, as previously explained in Section \ref{sec:ExpNTL}, the current system originally aimed to fully automatize the generation of campaigns, and the system-stakeholder interaction was low. Therefore, it would be necessary to modify the current system to allow the company's Stakeholder involvement in the learning process, allowing on-the-fly corrections of the model. 

\subsubsection{Our proposal}

\begin{figure}[h!]
    \centering
    \includegraphics[width=0.48\textwidth]{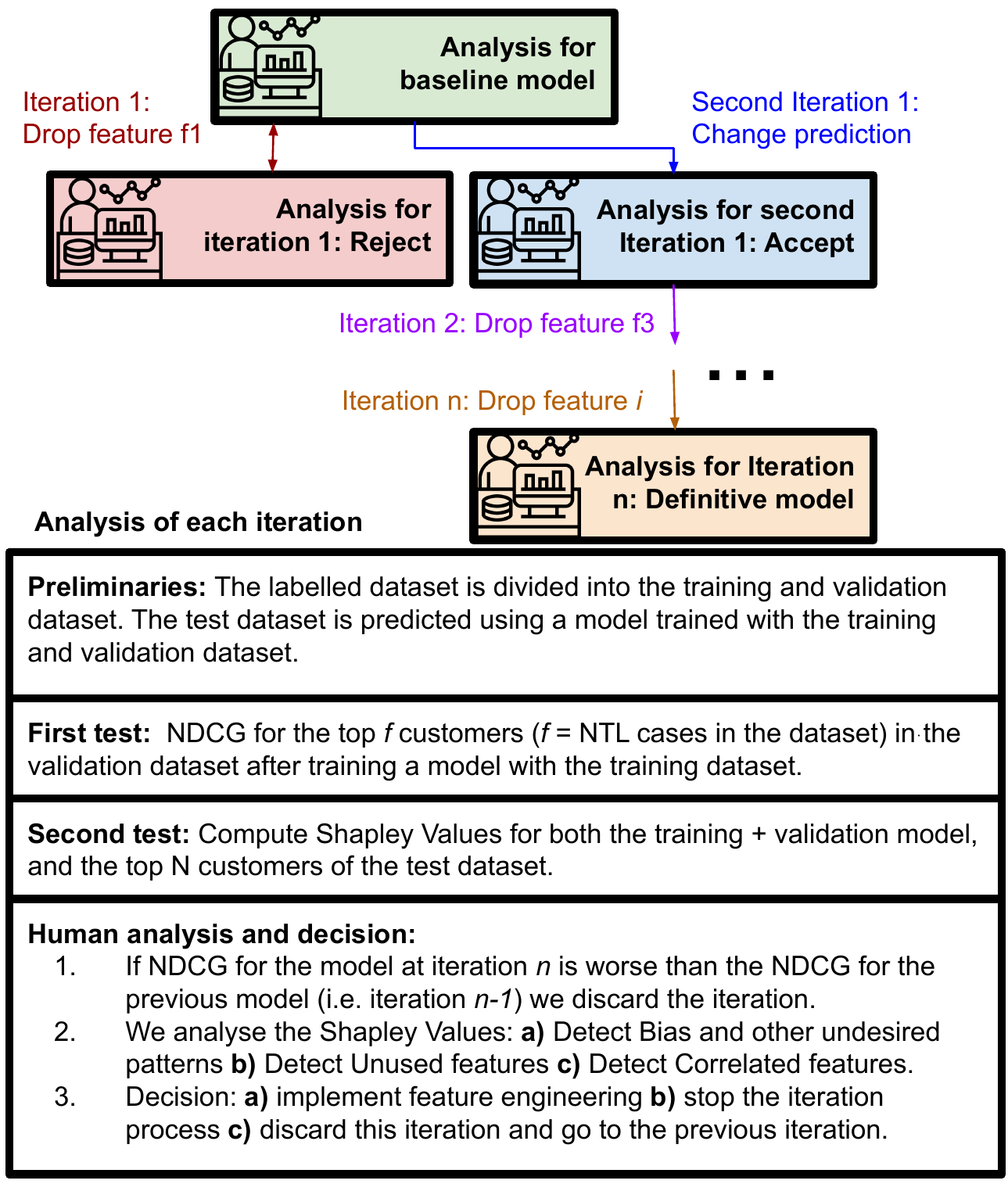}
    \caption{The building process is an iterative human-in-the-loop process where the stakeholder uses Shapley Values to guide the model's training process to achieve a more generalizable and fairer model.}
    \label{fig:cook}
\end{figure}

In this work we propose to involve the stakeholder through a human-in-the-loop solution to guide the system when training the model (Figure \ref{fig:cook}). In each iteration, the stakeholder analyzes through Shapley Values the patterns learned and implements feature engineering to correct biases and other data-related problems that are specific to that domain at the moment of building the campaign, as well as remove correlated or unused features to increase the system's interpretability, to achieve a simpler (i.e. with fewer variables), more understandable (i.e. with patterns validated by the stakeholder) and, therefore, a better model in terms of generalization. 

To benchmark each model we use the Normalized Discounted Cumulative Gain ($NDCG$, \cite{jarvelin2002cumulated}) to obtain a global vision of the quality of the predictions made by a model,
\begin{center}
    $NDCG_t =\dfrac{DCG_t}{iDCG_t} $
\end{center}

\noindent where $DCG_t$ is defined as

\begin{center}
$DCG_t =\dfrac{\sum_{i=1}^{t}{energy_{i}}-1}{log_2(i+1)} $

\end{center}

\noindent being $energy_i$ the amount of energy recovered in the visit made to the customer ranked at position $i$, and $iDCG$ corresponds to the maximum $DCG$ possible (i.e. a perfect prediction in terms of order). The $NDCG$ provides a generic vision of the correction of the model beyond any threshold. Other alternatives (e.g. $precision@k$, i.e. precision at the top $k$ instances) can tend the model to exploit the existing biases in the data and, therefore, the resulting model would not be generalizable.

\subsubsection{The Human Analysis in the Building Process}

With the information provided by the Shapley Values and the NDCG metric, the stakeholder has to analyze in each iteration $n$ the correctness of the model trained in comparison to the previous iteration $n-1$, and propose a new model in iteration $n+1$ by implementing feature engineering. This process is subjective (e.g. depending on the stakeholder there might be slight differences regarding what can be considered a good pattern), and also needs to be adapted to every domain (e.g. a good pattern in one domain might be a fair pattern in another domain). However, there are certain fundamentals that every analysis shares, summarized as follows, which we will use in this work: 

    \paragraph{The $NDCG$ should not decrease between iterations} In each iteration, we should not decrease the benchmarking performance on unseen data. Therefore, in each iteration, $NDCG_n >= NDCG_{n-1}$\footnote{We would accept some margin in this description, i.e. we consider that a model is worse in terms of $NDCG$ when the value is significantly lower (at least 0.1 lower).}.
    \paragraph{The outliers should be detected and processed} A Shapley Value from a high-scored instance that stands out in comparison to the rest of the Shapley Values can be a consequence of an outlier in the prediction labels (more specifically, an NTL case with a much higher value of kWh recovered than the rest of the NTL cases). In this case, the stakeholder should consider transforming the instance that causes the outlier to avoid biases on prediction.
    \paragraph{The system should be as simple as possible} To reduce the complexity of the model to increase generalizability on unseen data but also improve interpretability, we should remove features that have a low impact on the model. Also, we should remove correlated features with similar meanings that contribute similarly according to Shapley Values.
    
    

The correction of bias should have priority over removing a feature: a bias highly influences how a model is learned and, therefore, its correction can cause a feature with no importance in the biased model to gain relevance in the new model. All these considerations are explained in the short example that we provide in the following Section \ref{sec:CaseStudy}.


\section{A case study with a real dataset}
\label{sec:CaseStudy}

In this section, we exemplify the human-in-the-loop process and analyze the benefits of implementing it in our NTL system.

\subsection{Preliminaries}

\subsubsection{The Dataset used}

For the case study we use a real dataset\footnote{further information like the region and the typology of the customers is anonymized to protect the privacy of the data.} from the utility company with more than $1,000,000$ customers\footnote{The customers are apartments/small houses from the same Spanish region.}. The labelled instances include around 10,500 NTL cases, and almost 300,000 non-NTL cases and the dataset is split into three sub-datasets: the training (80\% of the labelled instances), the validation (10\% of the instances) and the test dataset (the remaining 10\%). Each partition is stratified. There is no timestamp consideration (i.e. we do not use the last 10\% of NTL cases as the test dataset) to guarantee diversity and reduce the differences between the datasets\footnote{If the stakeholder decides to visit recidivist customer in July and August, and in September, we split the data considering the timestamp, in the test dataset we would have an over-representation of the recidivist customers.}. 

\subsubsection{The Algorithm, Loss Function and Metric Used} 
\begin{figure}[h!]
    \centering
    \includegraphics[width=0.45\textwidth]{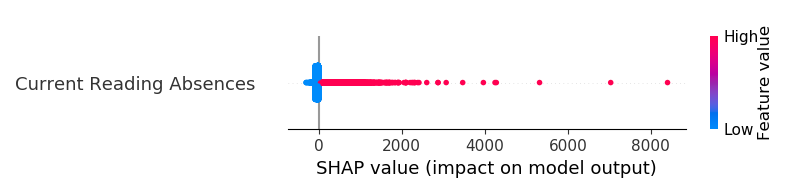}
    \caption{In red there are the high values of the features and, in blue, the low values. In this specific case we can see that having a high value in \emph{Current Reading Absences} (i.e. that the company has several months with no new meter readings) increases the $\hat{y}$ value of the instance.}
   \label{fig:exampleSHAP}
\end{figure}

The Gradient Boosting Model trained is a Root Mean Square Error Catboost Regressor, i.e. we consider the problem of detecting NTL as a point-wise ranking problem where we predict the amount of energy to recover for each customer. The methods used to analyze the correctness of our model are the $energy_{200}$ (to compare the energy recovered before and after the human-in-the-loop process), $NDCG$ on the validation dataset, and Shapley Values plots to analyze the patterns learnt by each model, as we show in Figure \ref{fig:exampleSHAP}. 

\subsubsection{Semantic Grouping of Features and Evaluation}

To facilitate the explanation and readability of this work, we exemplify the human-in-the-loop approach only on the visit-related features, including plots for the Shapley Values in the training and test dataset. A brief description of these features is the following:
\paragraph{Types of visits} Most of the visit-related variables represent the visits made to the customers and their three possible results. More specifically, the \emph{Fraud} features refer to the visits in which the company detected an NTL case. The \emph{Correct} features refer to the visits where the installation was checked, but no NTL was detected. The \emph{Impossible} features profile the visits with no conclusive result (in general, because the meter was not accessible). Finally, the \emph{Visit} features represent all the visits without NTL/Non-NTL distinction. Based on this information, we profile the following features:
\begin{itemize}
    \item \textit{Number of occurrences}: Those features that include the \# prefix refers to the occurrences of that type of visit  (e.g. \emph{\#Visit} refers to the number of visits the company has made to the customer).
    \item \textit{Last occurrence}: The last occurrence of each type of visit is represented with the features with the \textit{last} prefix (e.g. \emph{LastVisit} would refer to how many months have passed since the last visit). When the customer has never been visited, the value of the feature is empty\footnote{That is, there is no value assigned, i.e. a missing value, that the Catboost library is able to process. This solution is applied in all the features to represent the non-existence of a value, e.g. the non-existence of a visit for that customer.}. 
    \item \textit{Type of visit}: A visit to a customer is often prompted by suspicion of fraud. In other cases, the visit is related to more generic reasons (e.g. a generic revision of the meter). Both cases are reflected with suffix $1$ and $2$, respectively: \emph{LastFraud1} refers to how many months have passed since the last fraud was detected in which the reason to visit was a suspicion of fraud (or NTL), \emph{LastFraud2} corresponds to how many months have passed since the last fraud in which the reason to visit was not NTL-related. \emph{LastFraud} (with no suffix) corresponds to the features that groups both types of features. 
\end{itemize}
\paragraph{Region-related features} There are also features related to the density of fraud \textit{around} the customer. That is, \emph{\#FraudZone} indicates the historical number of NTL cases in a customer's zone\footnote{A zone corresponds to a technical term regarding the distribution of the electricity: nearby towns or neighborhoods in a big city share a zone.}. Similarly, \emph{\#FraudStreet} is the same information than the \emph{\#FraudZone} but focused specifically on the street where the customer lives, and \emph{\#FraudInBuilding} counts the historical NTL cases in the building where the customer lives. There exist for each feature a derivative (with a suffix \textit{1Year}) in which the information is bounded in the last year (e.g. \emph{\#FraudZone1Year} indicates the number of fraud cases in the region during the last 12 months).
\paragraph{Threats} There is a third group of features (\textit{\#threats} and \textit{LastThreat}) that refers to the threats of the customer to the technician, i.e. if the customer violently prevents the installation revision from being carried out. 
\paragraph{Energy Cut} Finally, the \textit{EnergyCut} feature indicates how many months have elapsed since the last energy cut by the company due to non-payment.

\subsection{Tests}

In this section we exemplify the process of stakeholder-system interaction by implementing the following: removing a feature due to its irrelevance, removing a correlated feature and correcting an outlier. We compare the baseline model and the resulting model in terms of $energy_{200}$ to see if, in addition to the improvement in terms of interpretability and bias reduction (that would help to increase the robustness in real campaigns), the resulting model also recovers more energy in the test dataset.

\subsubsection{\textbf{First Model (baseline)}}   
\label{sec:firstIt}

\begin{figure}[h!]
\centering
    \includegraphics[width=0.45\textwidth]{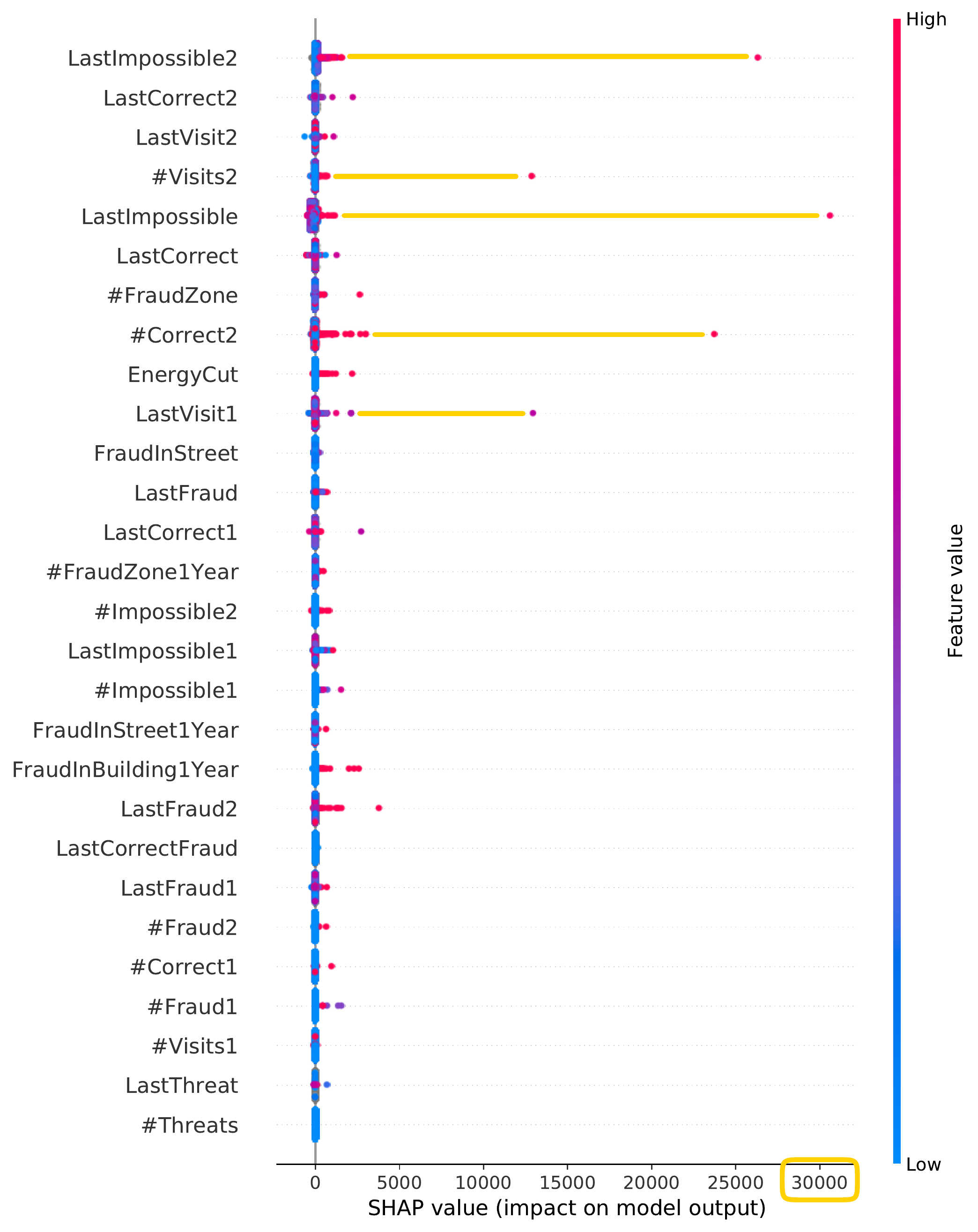}
    \caption{The outliers seen in the image (in yellow) are a consequence of an NTL case in which the amount of energy to recover is higher than 250000kWh. In this situation, the stakeholder in charge of the model building would consider reducing this prediction value to build a more unbiased model.}
    \label{fig:firstIt}
    
\end{figure}

\begin{itemize}
    \item $NDCG$: 0.44 in the validation dataset.
    \item $energy_{200}$: 249242.9kWh.
    \item Shapley Values: Figure \ref{fig:firstIt} (training+validation model).
\end{itemize}

\paragraph{Analysis} As we can see in Figure \ref{fig:firstIt}, our baseline model has Shapley Values that are abnormal, because the impact on the output is remarkably higher than all the other values for those features. For instance, if we analyze the \textit{LastImpossible2} feature, there is no compelling reason to justify that a feature value increases the output of the model up to 25,000kWh, while the second highest Shapley Value increases ten times less. Thus, this is an indicator of an outlier in the labelled information, i.e. an NTL case in which the company recovered a large amount of energy. In this case, the outlier corresponds to an NTL case in which the company recovered 260,000kWh, an extremely abnormal case of NTL due to the large amount of energy recovered\footnote{In the second NTL case in the dataset the company recovered around 50,000kWh. The typical customer consumption is close to 3,500kWh per year.}. With this information, the stakeholder would have two options: maintaining the outlier or correcting it. Maintaining an outlier could be useful in some specific cases (for instance, if the company aims to exploit biased patterns learnt\footnote{In some cases a biased pattern might be in line with business-related decisions. For instance, the stakeholder might consider not removing a pattern in which the customers from a region have higher predictions if the company aims to visit these type of customers more.}) but, in general, the stakeholder should consider its correction. 
\paragraph{Next step} In this case, an optimal solution would be reducing the weight of this NTL by modifying the label (for instance) four times (i.e. from 260,000 to 66,000kWh). With this change, we still indicate to the system that it is the higher NTL case in the labelled information, but we will avoid biases in the system.

\subsubsection{\textbf{Second Model (First iteration)}}

\begin{figure}[b!]
    \centerline{
    \includegraphics[width=0.45\textwidth]{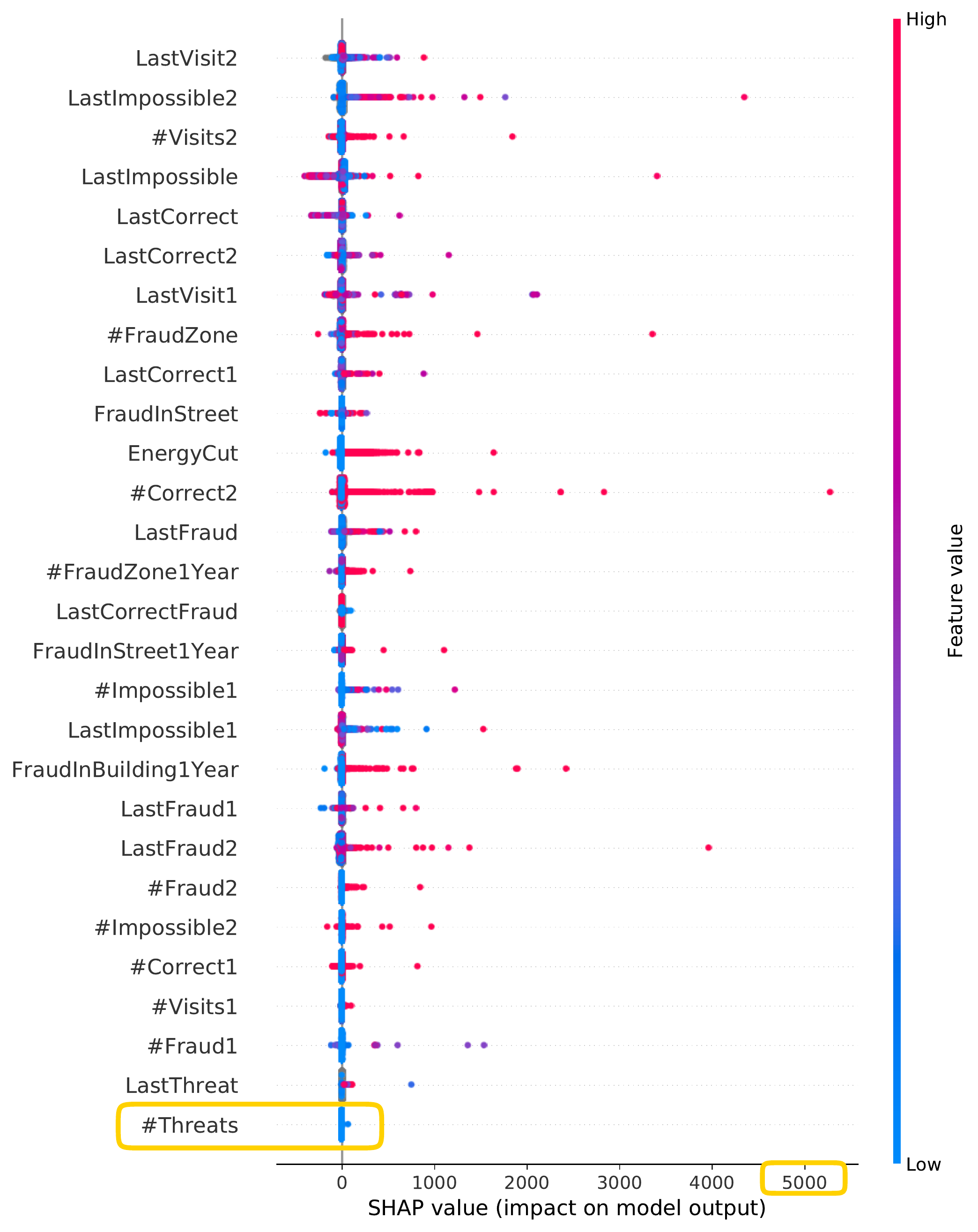}
    }
    \caption{The Shapley Values for the trained model indicates the non-relevance of the \emph{\#Threats} feature. Therefore, to facilitate the interpretation of the model by the stakeholders, we drop this feature from the training process.}
    \label{fig:secondIt}
    
\end{figure}

\begin{itemize}
    \item $NDCG$: 0.43 in the validation dataset.
    \item Shapley Values: Figure \ref{fig:secondIt} (training+validation model).
\end{itemize}

\paragraph{Analysis} First of all, we can see that we achieve a similar $NDCG$ value in the validation dataset, i.e. it seems that the unbiasedness does not reduce the prediction capacity of our model. Then, the Shapley Values from Figure \ref{fig:secondIt} seem to indicate that the model learnt is better: there are no outliers (the higher Shapley value is reduced from around 30000 to 5000), and therefore it should generalize better on unseen data. So, in summary, a stakeholder would prefer this model over the previous one.

\paragraph{Next step} For the next iteration we opt to drop the less important feature in the model: \emph{\#Threats}. This should not modify the model trained but would simplify the explanation provided to the stakeholders.

\subsubsection{\textbf{Third Model (Second Iteration)}}

\begin{figure}[h!]
    \centerline{
    \includegraphics[width=0.45\textwidth]{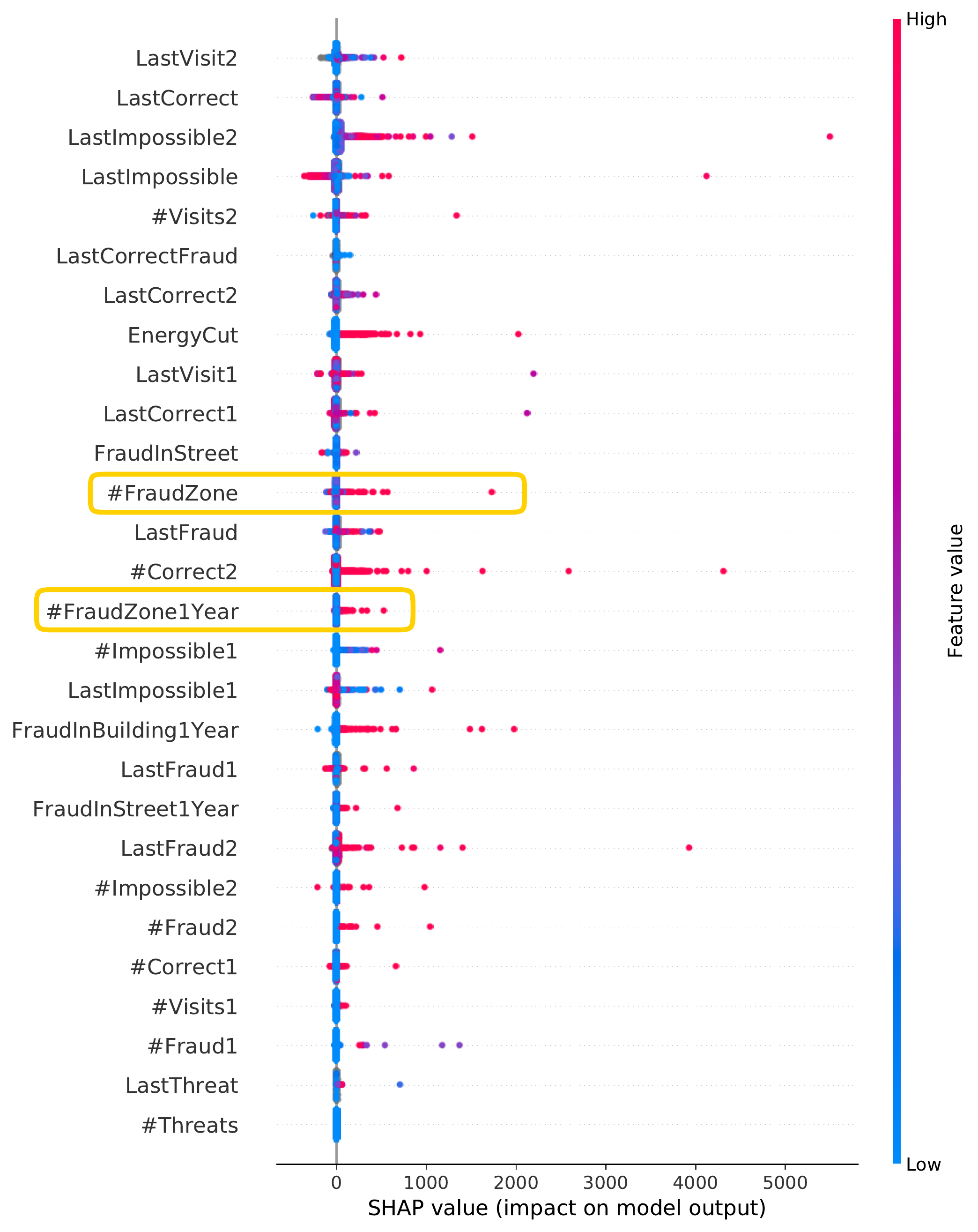}
    }
    \caption{The Shapley Values highlighted indicate that the model has learned similar patterns from both features with similar meanings. Removing one of the features would increase interpretability and reduce the curse of dimensionality. To decide the best feature to be removed, we can analyze how these patterns translate on unseen data (Figure \ref{fig:thirdIt_2}).}
    \label{fig:thirdIt_1}
    
\end{figure}

\begin{figure}[htpb!]
    \centering
     \includegraphics[width=0.45\textwidth]{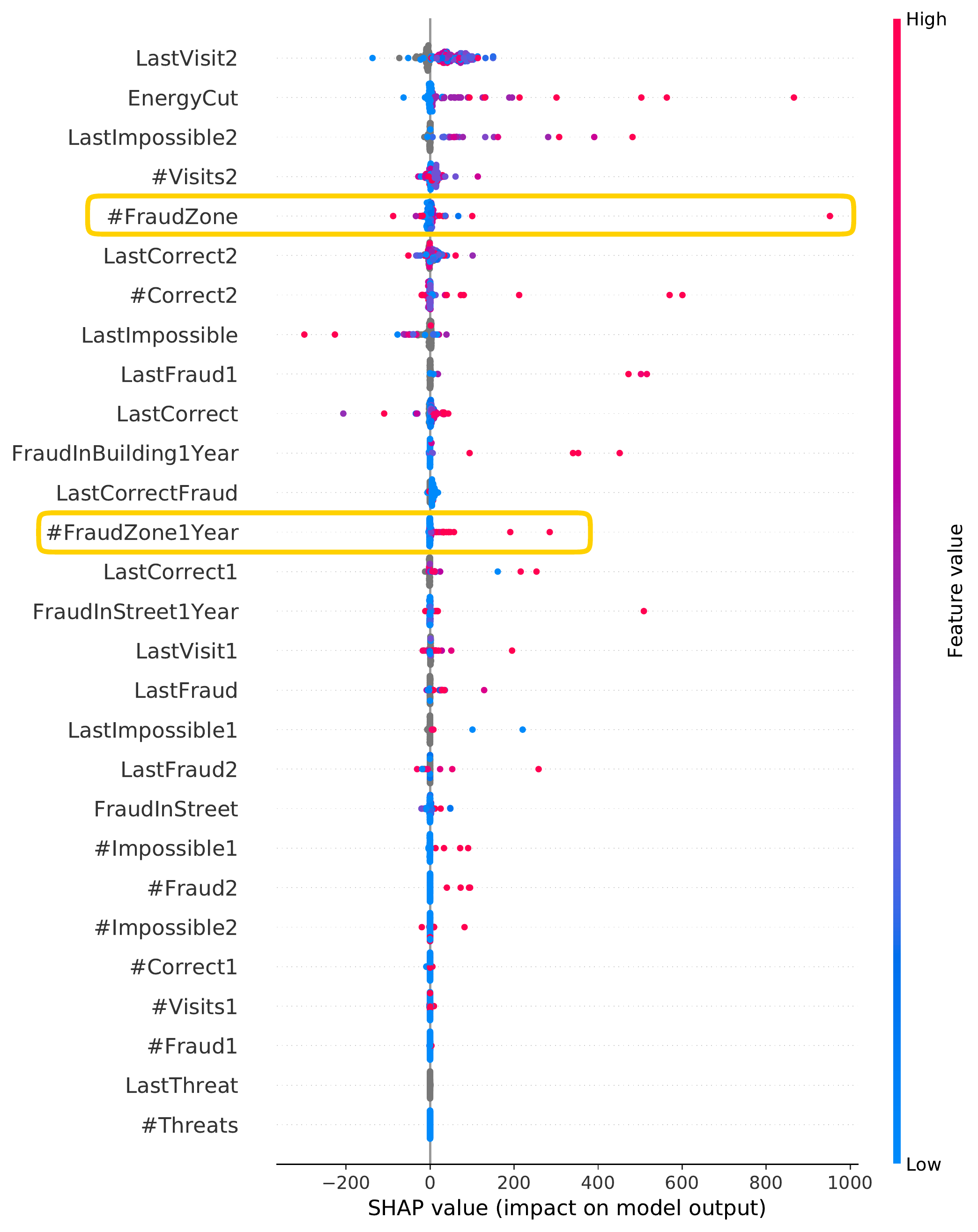}
    
    \caption{The Shapley Values in the top-scored 200 customers test dataset we see that how \emph{\#FraudZone1Year} influenced the prediction is much clearer than in the \emph{\#FraudZone} feature and, therefore, we would drop the latter feature from the dataset.}
    \label{fig:thirdIt_2}
    
\end{figure}

\begin{itemize}
    \item $NDCG$: 0.42 in the validation dataset.
    \item Shapley Values: Figures \ref{fig:thirdIt_1} (training+validation model) and \ref{fig:thirdIt_2} (top-scored customers from the test dataset).
\end{itemize}

\paragraph{Analysis} Dropping the \emph{\#threats} has not changed much, as expected, what the model has learned (i.e. the plot from Figure \ref{fig:secondIt} and the left plot from Figure \ref{fig:thirdIt_1} are similar). However, the possibility of dropping features can be fruitful from the company's perspective. First, it allows to correct undesired patterns learned that, from the human perspective, have no logic but can be seen in a biased dataset. When we introduce a feature in the system, we expect that the system will learn some specific patterns. For instance, when we profile with a feature that the customer is consuming much less than the average, we consider that the system should see this as an indicator of NTL. Therefore, if the system learns otherwise in one specific domain\footnote{If this undesired pattern is constantly learned in all the domains, then the feature drop would be definitive.}, the stakeholder can consider it appropriate to remove it in that specific campaign. Moreover, learning from fewer features with low relevancy helps avoid overfitting and increases the generalizability and interpretability of the model.

\paragraph{Next step} For this third iteration, we exemplify the process of removing a correlated feature from the model. As shown in Figure \ref{fig:thirdIt_1}, the features \emph{\#FraudZone} and \emph{\#FraudZone1Year} provide similar information to the learning process globally: a high number of NTL cases in the zone is an indicator of NTL. If we focus on the Shapley Values from the top-scored 200 customers, we can see that the patterns learnt from the \emph{\#FraudZone} feature are unclear\footnote{From the stakeholder's point of view, it is simpler to explain the \emph{\#FraudZone1Year} pattern "high values is an indicator of NTL" than the patterns from \emph{\#FraudZone}, which are unclear, where sometimes a high value has positive Shapley Values,  and in other cases, it has negative Shapley Values.} and, for this case, we would opt to remove the \emph{\#FraudZone} feature.

\subsubsection*{Resulting Model}

The resulting model corresponds to the baseline model + correction of the bias + \emph{\#threats} drop due to its low relevance + \emph{\#FraudZone} drop (correlated with \emph{\#FraudZone1Year}).

\begin{itemize}
    \item $NDCG$: 0.44
    \item $energy_{200}$: 257038.7kWh
\end{itemize}

\paragraph{Analysis} The resulting model, in terms of $NDCG$, is as good as the vanilla model, and in terms of $energy_{200}$ is slightly better, recovering around 8000kWh more energy. However, in terms of Shapley Values the resulting model is more trustworthy from the stakeholder's point of view, and should generalize better on unseen data. 

This example is rather naive since we have only slightly modified the system by implementing feature engineering. However, it exemplifies the benefits of the human-in-the-middle approach in which the stakeholder guides the system to learn an optimal model, mitigating the specific biases and other problems regarding the use of observational data. Moreover, the fact that the stakeholder is an active part of the system has positive consequences beyond the ones mentioned above (i.e. the better generalization on unseen data and the better interpretability), as the company can trust the system much more, one of the objectives of explainable AI \cite{arrieta2020explainable}.

\section{Technical Considerations, Conclusions and Future Work}
\label{sec:Conclusions}

Building an NTL detection system using observational data is challenging due to the existence of biases and other data-related problems. This is aggravated when the predictive model is a black-box algorithm due to its opacity, making it difficult to implement solutions to mitigate these problems. We exemplify this problem in the NTL detection system we have been developing for a utility company from Spain, and propose a human-in-the-loop approach to increase the Accessibility, the Interactivity and the Informativeness of our NTL detection system. Moreover, the stakeholder's guidance guarantees the causality of the learned models: nonetheless, in certain cases it might be difficult to discern a good correlation and a real causal pattern, the existence of a human oracle avoids the existence of bad correlations and undesired patterns. 

This approach can be easily implemented in any similar NTL detection approach. However, according to our experience, using a GPU-accelerated state-of-the-art Gradient Boosting (GBM) library and the TreeSHAP \cite{lundberg2018consistent} implementation is the optimal approach since it provides fast, explained and accurate out-of-the-box predictions. There exist other GPU accelerated predictive algorithms (e.g. Deep Learning or Support Vector Machines) that might also provide accurate results but need either more data processing and the explanation approaches are slower (e.g. the KernelSHAP from SHAP) or the explanation obtained is less detailed (e.g. Feature Importance).

Future work would focus on two aspects. In the short term, our effort will focus on improving this system-stakeholder interaction based on the stakeholder's feedback. In the long term we will explore if the system can robustly assist the stakeholder by suggesting the modifications needed to achieve more robust models or directly if the process can be automatised with an expert system.

\section*{Acknowledgements}
This work has been supported by MINECO and FEDER funds under grant TIN2017-86727-C2-1-R, and a collaboration with Naturgy.



%
\bibliographystyle{IEEEtran}
\bibliography{doc.bib}

\end{document}